\documentclass{article}
\usepackage{hello}

\usepackage[margin=1in]{geometry}

\usepackage{graphicx}
\usepackage{booktabs}
\usepackage{array}
\usepackage{amsmath}
\usepackage{amssymb}
\usepackage{amsfonts}
\usepackage{multirow}
\usepackage{verbatim}
\usepackage{caption}
\usepackage{longtable}
\usepackage{supertabular}

\usepackage{enumitem}
\usepackage{tablefootnote}
\usepackage[round,semicolon]{natbib}
\usepackage{colortbl}
\usepackage{xspace}
\usepackage{textcomp}
\usepackage{makecell}
\usepackage{multirow}
\usepackage{lscape} 
\usepackage{siunitx}

\usepackage{enumitem}
\usepackage{tablefootnote}

\usepackage{colortbl}
\usepackage{xspace}
\usepackage{textcomp}
\usepackage{makecell}
\usepackage{multirow}
\usepackage{lscape} 
\usepackage{siunitx}
\definecolor{dt}{gray}{0.7}
\usepackage{pifont}       
\usepackage{bbding}       
\usepackage{fontawesome}

\usepackage{amssymb}
\usepackage{pifont}
\usepackage{scrextend}

\usepackage{array}
\usepackage{tgpagella}
\usepackage{latexsym}
\usepackage[T1]{fontenc}
\usepackage[utf8]{inputenc}
\usepackage{microtype}
\definecolor{mydarkblue}{rgb}{0,0.08,0.45}
\usepackage[colorlinks,citecolor=mydarkblue,urlcolor=mydarkblue,linkcolor=mydarkblue]{hyperref}
\usepackage{url}            
\usepackage{nicefrac}       
\usepackage{changepage}
\usepackage{xargs}          
\usepackage{wrapfig,lipsum,booktabs}
\usepackage{longtable}
\usepackage{subcaption}
\usepackage{endnotes}

\usepackage{pgfplots}
\usetikzlibrary{pgfplots.groupplots}
\pgfplotsset{compat=1.3}
\usepackage{tikz}
\usetikzlibrary{patterns}

\usepackage[most]{tcolorbox}

\usepackage[capitalize,noabbrev]{cleveref}
\crefname{section}{Section}{\S\S}
\Crefname{section}{Section}{\S\S}
\crefname{table}{Table}{Tables}
\crefname{figure}{Figure}{Figures}
\crefname{algorithm}{Algorithm}{}
\crefname{equation}{eq.}{}
\crefname{appendix}{Appendix}{}
\crefformat{section}{Section #2#1#3}
\usepackage{multicol}
\usepackage{multirow}

\usepackage{titlesec}
\titleformat*{\section}{\large\bfseries}

\definecolor{battleshipgrey}{rgb}{0.3, 0.3, 0.3}
\definecolor{brilliantrose}{rgb}{1.0, 0.33, 0.64}
\definecolor{americanrose}{rgb}{1.0, 0.01, 0.24}
\definecolor{jweigreen}{rgb}{0,0.45,0.24}
\definecolor{bluegray}{rgb}{0.1, 0.1, 0.4}
\definecolor{ao(english)}{rgb}{0.0, 0.5, 0.0}
\definecolor{blanchedalmond}{rgb}{1.0, 0.92, 0.8}
\definecolor{atomictangerine}{rgb}{1.0, 0.6, 0.4}
\definecolor{chocolate(web)}{rgb}{0.82, 0.41, 0.12}
\definecolor{bananayellow}{rgb}{1.0, 0.88, 0.21}
\definecolor{goldenbrown}{rgb}{0.6, 0.4, 0.08}
\definecolor{aliceblue}{rgb}{0.94, 0.97, 1.0}
\definecolor{beige}{rgb}{0.96, 0.96, 0.86}
\definecolor{babyblue}{rgb}{0.54, 0.81, 0.94}
\definecolor{camel}{rgb}{0.76, 0.6, 0.42}
\definecolor{cinnamon}{rgb}{0.82, 0.41, 0.12}
\definecolor{deepskyblue}{rgb}{0.0, 0.75, 1.0}
\definecolor{frenchblue}{rgb}{0.0, 0.45, 0.73}
\definecolor{classicrose}{rgb}{0.98, 0.8, 0.91}
\definecolor{frenchrose}{rgb}{0.96, 0.29, 0.54}
\definecolor{frenchlilac}{rgb}{0.53, 0.38, 0.56}
\definecolor{frenchbeige}{rgb}{0.65, 0.48, 0.36}
\definecolor{verylightgreen}{RGB}{240, 255, 235}
\definecolor{verylightred}{RGB}{255, 235, 235}
\definecolor{verylightyellow}{RGB}{255, 254, 235}
\definecolor{dt}{gray}{0.7}

\definecolor{forestgreen}{HTML}{2e7d43}
\definecolor{color1}{HTML}{FF9999}
\definecolor{color2}{HTML}{FF6666}
\definecolor{color3}{HTML}{FF3333}
\definecolor{color4}{HTML}{E60000}
\definecolor{color5}{HTML}{B30000}
\definecolor{color6}{HTML}{8CD98C}
\definecolor{color7}{HTML}{53c653}
\definecolor{color8}{HTML}{39ac39}
\definecolor{color9}{HTML}{2d862d}
\definecolor{color10}{HTML}{206020}
\definecolor{color11}{HTML}{cca300}

\usepackage{minitoc}
\usepackage{float}

\newlength\savewidth
\newcommand{\tablestyle}[2]{\setlength{\tabcolsep}{#1}\renewcommand{\arraystretch}{#2}\centering\footnotesize}

\usepackage{amsmath,amsfonts,bm}

\def\eqref#1{equation~\ref{#1}}

\def\1{\bm{1}}

\DeclareMathAlphabet{\mathsfit}{\encodingdefault}{\sfdefault}{m}{sl}
\SetMathAlphabet{\mathsfit}{bold}{\encodingdefault}{\sfdefault}{bx}{n}

\title{
\textbf{
EVLM: An Efficient Vision-Language Model for Visual Understanding}
}

\author{
\large{}
Kaibing Chen \hspace{6mm} Dong Shen \hspace{6mm} Hanwen Zhong \hspace{6mm} Huasong Zhong \hspace{6mm} Kui Xia \\
Di Xu \hspace{6mm} Wei Yuan \hspace{6mm} Yifei Hu \hspace{6mm} Bin Wen \hspace{6mm} Tianke Zhang \hspace{6mm} Changyi Liu \\
Dewen Fan \hspace{6mm} Huihui Xiao \hspace{6mm} Jiahong Wu \hspace{6mm} Fan Yang$^{\dag}$ \hspace{6mm} Size Li \hspace{6mm} Di Zhang
\\
\\
\large{}
Kuaishou Technology
}

\date{}

\begin{document}

\doparttoc 
\faketableofcontents 

\maketitle

\begin{abstract}
\noindent
In the field of multi-modal language models, the majority of methods are built on an architecture similar to LLaVA. These models use a single-layer ViT feature as a visual prompt, directly feeding it into the language models alongside textual tokens. However, when dealing with long sequences of visual signals or inputs such as videos, the self-attention mechanism of language models can lead to significant computational overhead. Additionally, using single-layer ViT features makes it challenging for large language models to perceive visual signals fully.
This paper proposes an efficient multi-modal language model to minimize computational costs while enabling the model to perceive visual signals as comprehensively as possible. Our method primarily includes: (1) employing cross-attention to image-text interaction similar to Flamingo. (2) utilize hierarchical ViT features. (3) introduce the Mixture of Experts (MoE) mechanism to enhance model effectiveness. Our model achieves competitive scores on public multi-modal benchmarks and performs well in tasks such as image captioning and video captioning.
\end{abstract}

{\let\thefootnote\relax\footnotetext{$^\dag$Corresponding author}}



\begin{figure*}[t]
\centering
\includegraphics[width= 1\textwidth]{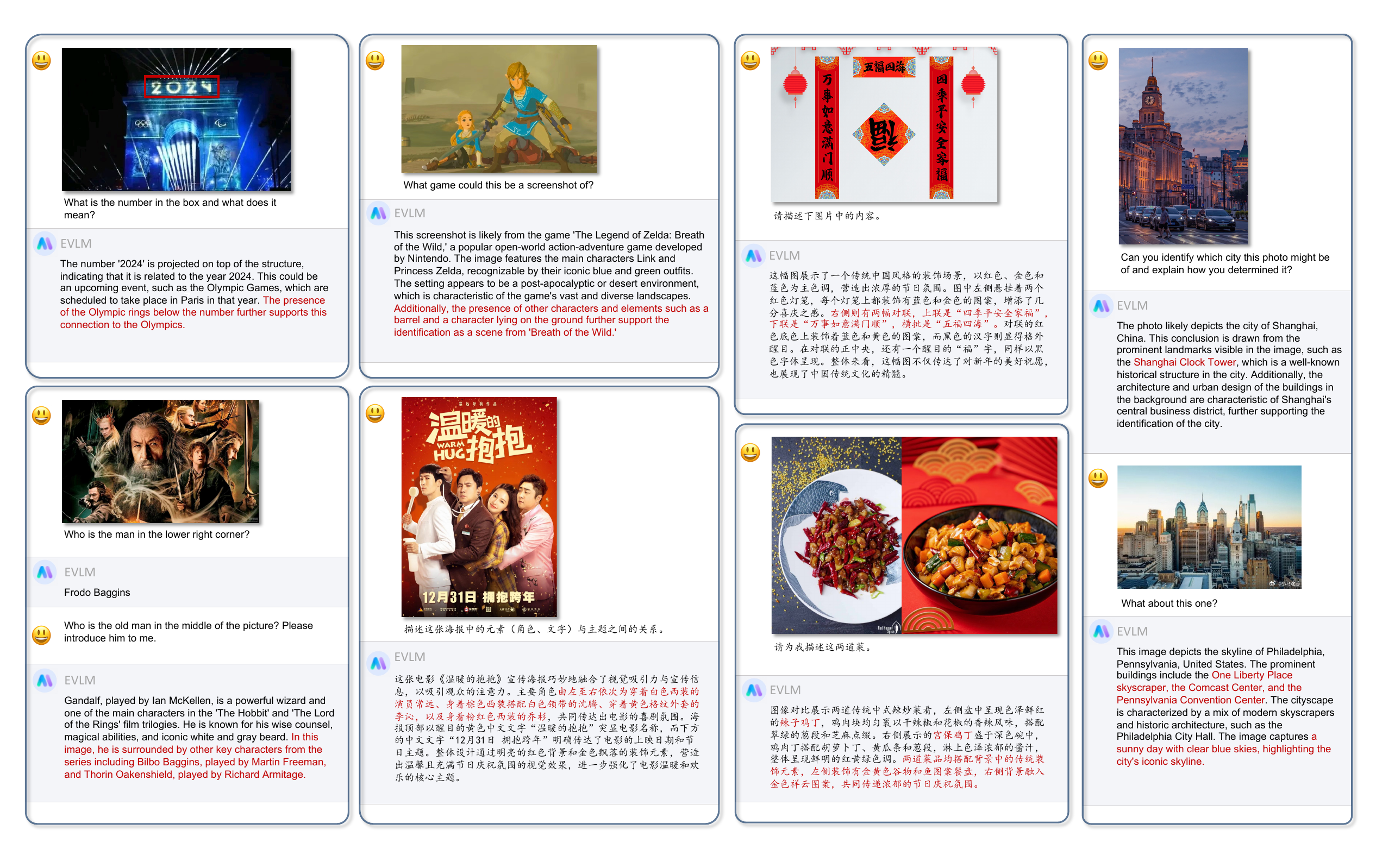}
\caption{Some qualitative examples generated by our model.}
\label{example}
\end{figure*}

\section{Introduction}

Recently, both academia and industry have seen the emergence of numerous outstanding large language models~\citep{gpt3,achiam2023gpt,anil2023palm,gao2023llama,team2023gemini,bai2023qwen,team2023internlm,zeng2022glm,young2024yi}, these models not only reduce the gap to the performance of GPT-4 but also excel across significant benchmarks. These powerful language models have fostered the development of vision-language models. Today's vision-language models can handle various visual tasks, including object recognition, object localization, OCR (optical character recognition), and document analysis. These advancements have significantly improved the model's ability to process complex visual information and generate accurate textual descriptions.

Researchers have explored many methods to enhance the perceptual capabilities of vision-language models for details. Some studies~\citep{lin2023sphinx,jain2024vcoder,hong2024cogagent,lee2024collavo} focus on using multiple visual encoders to enhance the encoding of visual signals, while others~\citep{liu2024llava,li2024mini,liu2024infimm,chen2024far,dong2024internlm} aim to improve the model's perception of small objects by increasing image resolution or slicing high-resolution images. These methods have notably boosted the performance of vision-language models in detail recognition, particularly in tasks such as OCR and document analysis. Once visual signals are encoded into features, the resulting features often require long token lengths. Attempts to use multiple encoders to extract visual features, increase input image resolution, or slice images, especially when dealing with video data or multiple image inputs, these operations significantly increase the length of visual tokens. In models like LLaVA~\citep{liu2023visual}, directly feeding excessively long visual tokens into language models will result in substantial computational overhead. On the other hand, adopting approaches akin to Q-former~\cite {blip2} for compressing visual features may lead to loss of visual information. Therefore, choosing appropriate strategies to balance computational efficiency and information richness is crucial in designing efficient vision-language models.

This paper proposes an efficient visual-language model that adopts a cross-attention mechanism similar to Flamingo~\citep{alayrac2022flamingo} for interaction between visual and textual inputs. Adopting cross-attention ensures that even with long visual tokens, controlling the feature dimensions in cross-attention does not lead to excessive computational overhead. To feed sufficient visual features into the language model, hierarchical ViT features are employed, enabling the large-scale language model to perceive visual signals at different levels, thus aiding in understanding tasks of varying granularity. Additionally, to enhance model performance, the Mixture of Experts (MoE) is applied on the Cross Attention to scale trainable model parameters. Extensive pre-training on a large-scale dataset of bilingual image-text pairs enables our visual-language model to acquire rich visual-linguistic knowledge. Leveraging our pre-trained model and refined visual feature input design, our model achieves competitive scores on public multimodal benchmarks and demonstrates exemplary performance in tasks such as image and video captioning. Fig.~\ref{example} shows some qualitative examples generated by our model.

\section{Model Architecture}

\begin{figure*}[ht]
\centering
\includegraphics[width= 1\textwidth]{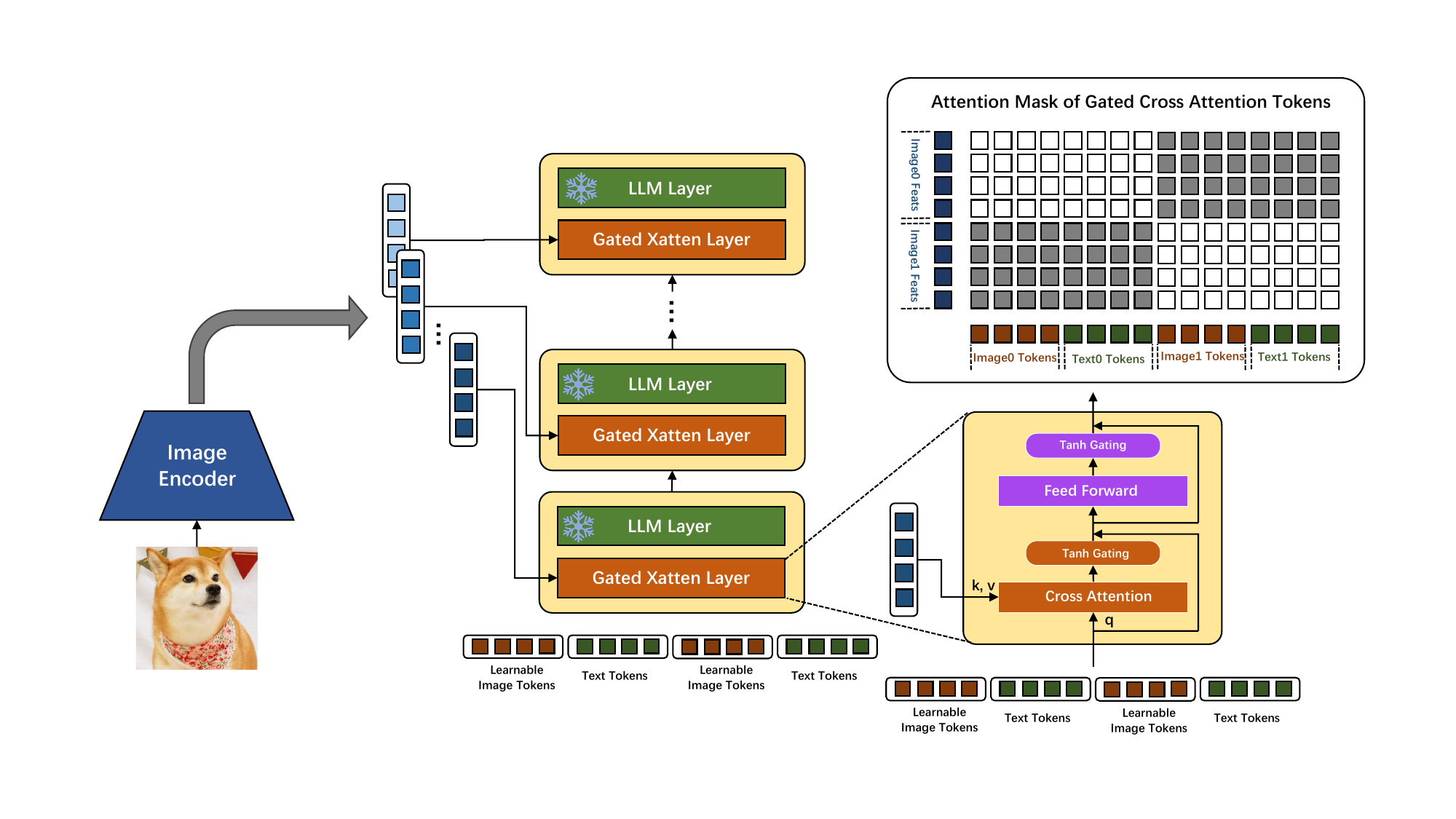}
    \caption{The framework diagram of our multi-modal model.}
\label{mllm_framework}
\end{figure*}

Our model architecture is based on Flamingo~\citep{alayrac2022flamingo}, primarily consisting of a visual encoder, a large language model, and a Gated Cross Attention Layer. To enable the multi-modal model to capture more fine-grained visual signals, we extracted hierarchical visual features from different layers of the visual encoder and increased the length of Flamingo's media tokens. Fig.~\ref{mllm_framework} is our model framework diagram.

\textbf{Visual Encoder}: To enhance our multi-modal model's visual perception capability, we utilized the 4.4B EVA2-CLIP-E-Plus~\citep{sun2023eva} model. In practice, we removed the norm and head layers after the last transformer block. To extract hierarchical visual features, we uniformly sampled 8 feature sequences from the last 40 layers of the transformer and sequentially fed these 8 feature sequences into different Gated Cross Attention layers of Flamingo.

\textbf{Gated Cross-Attetion Layer}: Similar to Flamingo, we use gated cross-attention to interact between vision and text. Unlike Flamingo, we replace the media token <image> with a set of learnable tokens of sequence length 16, hoping these learnable tokens can carry visual features similar to Qformer. Because not all text sequences are necessarily related to visual features, we pad a set of all-zero vectors on the visual feature sequence. The attention mask for learnable tokens, text sequences, and visual features is shown in Fig.~\ref{mllm_framework}, where each set of learnable tokens can only interact with the corresponding image, and text sequences can only interact with the previous image in the multi-modal sequence.

\textbf{Large Language Model}: We used the Qwen-14B-Chat 1.0 ~\citep{bai2023qwen}version of the language model, showing remarkable performance in content understanding and logical reasoning. To condition the language model on visual inputs, we insert a gated
cross-attention layer before every transformer layer of the language model.

\begin{figure*}[ht]
\centering
\includegraphics[width= 1\textwidth]{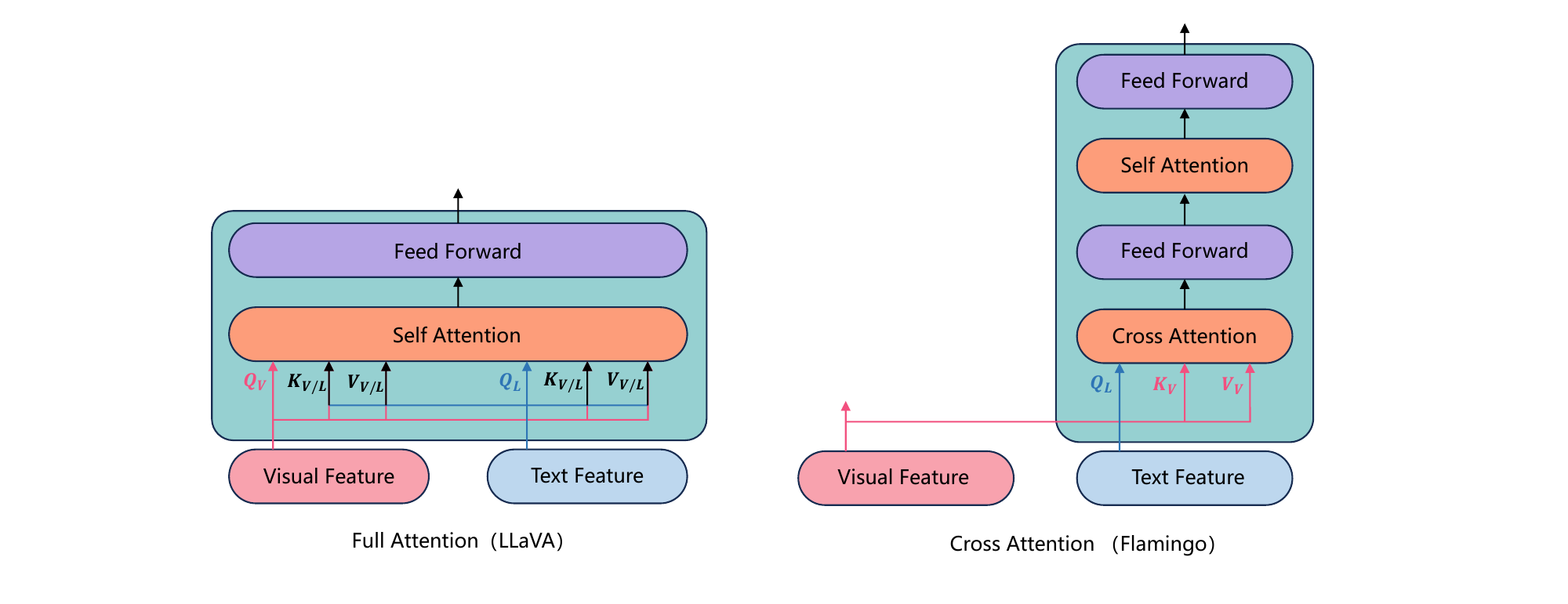}
    \caption{Full Attention and Cross Attention used in multi-modal model.}
\label{mllm_attention}
\end{figure*}

\textbf{Disscusion on Efficient Training}: In this section, we analyze the computing budget estimation of our EVLM and compare it with the result of current mainstream model architectures, such as the LLaVA family and Flamingo family. In the FLOPs estimation, we evaluate the attention and FFN layers within a single Transformer layer in the LLM.
As shown in \cref{mllm_attention}, $h_{\text{llm}}$ denotes the hidden state size of the LLM, while $d_{\text{img}}$ represents the dimension of visual representations. Moreover, in the Gated Cross-Attention Layer, the ratio of the attention layer to $h_{\text{llm}}$ is denoted as $r_{x_c}$, and the ratio of the FFN layer to $h_{\text{llm}}$ is also denoted as $r_{x_f}$. We distinguish between concatenation and cross-attention interaction modes, referred to as $\textbf{FLOPs}_{\text{full-attention}}$ and $\textbf{FLOPs}_{\text{cross-attention}}$, respectively. The total FLOPs can be estimated as follows:


\begin{equation}
\textbf{FLOPs}_{\text{full-attention}}=24B  (s_{\text{img}}+s_{\text{txt}})   h_{\text{llm}}^2+4B  (s_{\text{img}}+s_{\text{txt}})^2   h_{\text{llm}},
\label{eq:flops}
\end{equation}

\begin{align}
\textbf{FLOPs}_{\text{cross-attention}} = & 4(6+r_{x_c}+r_{x_f})B (16+s_{\text{txt}})   h_{\text{llm}}^2  + 4B (16+s_{\text{txt}})^2   h_{\text{llm}}  \nonumber \\
& + 4r_{x_c}  B   s_{\text{img}}   d_{\text{img}}    h_{\text{llm}}  + 4r_{x_c}  B   (16+s_{\text{txt}})   s_{\text{img}}   h_{\text{llm}},
\label{eq:flops}
\end{align}


Where $B$ denotes the batch size, $s_{\text{img}}$ and $s_{\text{txt}}$ denote the length of visual embeddings and text token sequences, respectively. The token sequence length is significantly reduced by employing a gated cross-attention layer, effectively lowering FLOPS and achieving efficient training. Specifically, with a length \( h \) of 5120 and \( d \) of 1792, and with \( r_{x_c} \) and \( r_{x_f} \) set to 0.2 and 0.5 respectively, we observed significant FLOPS reductions across various pre-training stages. FLOPS were reduced to \( S \) times the original, where $ \textbf{S} = \frac{\textbf{FLOPs}_{\text{cross-attention}}}{\textbf{FLOPs}_{\text{full-attention}}} $. For example, in multi-modal pre-training, \( s_{\text{img}} \) was 256 and \( s_{\text{txt}} \) was 64, yielding an \( \textbf{S}_{P} \) of 0.24. During continual pre-training, \( s_{\text{img}} \) was 1024 and \( s_{\text{txt}} \) was 64, resulting in an \( \textbf{S}_{CP} \) of 0.077. These results show a significant improvement in training efficiency.

\section{Training}

Our training process consists of three stages: multi-modal pre-training, multi-task continual pre-training, and multi-modal instruction fine-tuning.

\subsection{Multi-modal Pre-training}

Our multi-modal pre-training aims primarily at two objectives: 1) Cross-modal alignment of images and text, and 2) Modeling the intrinsic relationships within multi-modal data. We collected a large-scale dataset of image-text captions and web-type multi-modal data based on these objectives. For the image-text caption data, we implemented a data cleaning process to filter out anomalies such as images with unusual aspect ratios and text with repetitive words and to ensure relevance between images and text. We applied relevance filtering similar to MMC4 ~\citep{zhu2024multimodal} for web-type multi-modal data to retain highly correlated images. The detailed data processing procedures are documented in the appendix ~\ref{app:data-cleaning-processes}. The table~\ref{tab:pretraining_data} illustrates the distribution of our pre-training data. We obtained 2.5 billion image-text caption data and 50 million web-type multi-modal data. It is worth noting that 60\% of this data consists of Chinese, including a significant amount of self-built Chinese caption data. This was done to enhance the fine-grained alignment capability of our multi-modal model, covering specific visual concepts such as celebrity, landmark building, and dish.

\begin{table}[ht]
    \centering
    \caption{Details of Our pre-training data. LAION-en and LAION-zh are the English 
 and Chinese language subset of LAION-5B~\citep{laion5b}. LAION-COCO~\citep{laioncoco} is a synthetic dataset generated from LAION-en. DataComp~\citep{datacomp} and Coyo~\citep{coyo} are collections of image-text pairs. BLIP-cap is the bootstrapped pre-training datasets used by BLIP~\citep{blip}. MMC4~\citep{zhu2024multimodal} and WanJuan~\citep{he2023wanjuan} are the corpus of images interleaved with text.} 
    \tablestyle{6pt}{1.1}
    \begin{tabular}{ll ccc}
         \toprule
         \textbf{Language} & \textbf{Dataset} & \textbf{Type} & \textbf{Cleaned} \\
         \midrule
         \multirow{8}{*}{English} & BLIP-cap     & Caption       & 100M  \\
         & LAION-COCO   & Caption     & 40M   \\
         & LAION-en     & Caption     & 200M  \\
         & Coyo         & Caption     & 160M  \\
         & DataComp     & Caption     & 500M  \\
         & MMC4         & Web         & 40M   \\
         \midrule
         \multirow{2}{*}{Chinese} & LAION-zh     & Caption     & 100M  \\
         & \color{dt}In-house Data & \color{dt}Caption & \color{dt}1.4B  \\
         & WanJuan         & Web         & 10M \\
         \midrule
         & \multirow{2}{*}{Total}   & Caption     & 2.5B  \\
         & & Web                  & 50M  \\
         \bottomrule
    \end{tabular}
    \label{tab:pretraining_data}
\end{table}

During model training, we concatenated the caption and multi-modal web-type data separately to ensure each sample had up to 64 images and a sequence length of 2048, resulting in a total of 60 million training samples. In the first 25\% phase of training, only the parameters of the Gated Cross Attention Layer were trained. In the subsequent 75\% phase, we unfrozen the parameters of the latter half of the Visual Encoder for training. The input image size was $224 \times 224$ during this phase. The training objective was to minimize the cross-entropy of the text tokens. We employed a cosine learning rate strategy with a maximum learning rate of $6.4e^{-4}$. We completed training on the entire set of 60 million training samples. The detailed training hyperparameter settings are documented in the appendix ~\ref{app:hyperparam}.

\subsection{Multi-task Continual Pre-training}

We introduce the multi-task continual pre-training stage between the multi-modal pre-training and instruction fine-tuning. Compared with the pre-training stage, the continual pre-training stage pays more attention to MLM's high-level visual question-answering ability. Compared with the SFT stage, the continual pre-training stage is still about acquiring ability, not activating ability.

In the continual pre-training stage,  our training data sources are categorized into five distinct parts: Visual Question Answering (VQA) data, Natural Language Processing (NLP) data,  OCR data, detection data, and data which are sampled from the first pre-training stage to prevent catastrophic forgetting. The VQA data mainly comes from open-source data. The OCR and detection datasets combine open-source data and data generated through our simulations. The detailed data processing procedures of OCR are documented in the appendix ~\ref{app:ocr-data-cleaning-processes}. The NLP data is obtained from internal resources. Table~\ref{tab:multitask_data} shows the specific data proportions and sources. Finally, We create interleaved image-text data by packing the same task data into sequences of length 2048 and increasing the image resolution from $224\times 224$ to $448\times 448$.

In this phase, we unfrozen the parameters of the latter half of the Visual Encoder and gated cross-attention layer for training. The training objective was to minimize the cross-entropy of the text tokens. We employed a cosine learning rate strategy with a maximum learning rate of $1e^{-4}$. The model obtained at this stage is called EVLM-Base. The detailed training hyperparameter settings are documented in the appendix ~\ref{app:hyperparam}.

\begin{table}[ht]
    \centering
    \caption{Details of multi-task continual  pre-training data. 
    }
    \tablestyle{6pt}{1.1}
    \begin{tabular}{l c l}
         \toprule
         \textbf{Task} & \textbf{\# Samples} & \textbf{Dataset} \\
         \midrule
         partially sampled stage1 data      & 30M  & \makecell[l]{The data were clustered and then sampled according to their cluster IDs.  } \\
         VQA            & 9M  & \makecell[l]{GQA, VGQA, VQAv2, DVQA, OCR-VQA, DocVQA, \\ TextVQA, ChartQA, AI2D, mmicl, Simulation data} \\
         Detection & 17M  & GRIT, Visual Genome, RefCOCO, RefCOCO+, RefCOCOg \\
         OCR            & 26M & SynthDoG-en \& zh, Common Crawl pdf \& HTML, Simulation data \\
         nlp data & 10M & \color{dt}In-house Data \\
         \bottomrule
         total & 92M & \\
         \bottomrule
    \end{tabular}
    \label{tab:multitask_data}
\end{table}

\subsection{Supervised Fine-tuning}
\subsubsection{Dense Baseline Model}
During this stage, we finetuned our EVLM-Base through instruction finetuning to activate its instruction-following abilities. We used a broad range of high-quality instruction tuning data, totaling 2.3 M samples. As illustrated in the table, these include: 1) User Instruct Data: We incorporate the ShareGPT-4V and LLaVA-ZH datasets. 2) Multimodal Document/Chart Data: We used DocVQA and SynDog-EN to enhance the model's document comprehension capabilities. Following Qwen VL-7B Chat, we also used ChartQA, DVQA, and AI2D to understand charts and diagrams better. 3) Math Problems: We used MathInstruct, MathPlus, and geoqa+ data to improve the model's mathematical reasoning ability.

In this phase, we froze the LLM and tuned only the cross-attention layers and the last quarter ViT layers, achieving robust performance. The model obtained at this stage is called EVLM-Chat.

\subsubsection{Scaling via Mixture-of-Experts}
\begin{figure*}[ht]
\centering
\includegraphics[width= 1\textwidth]{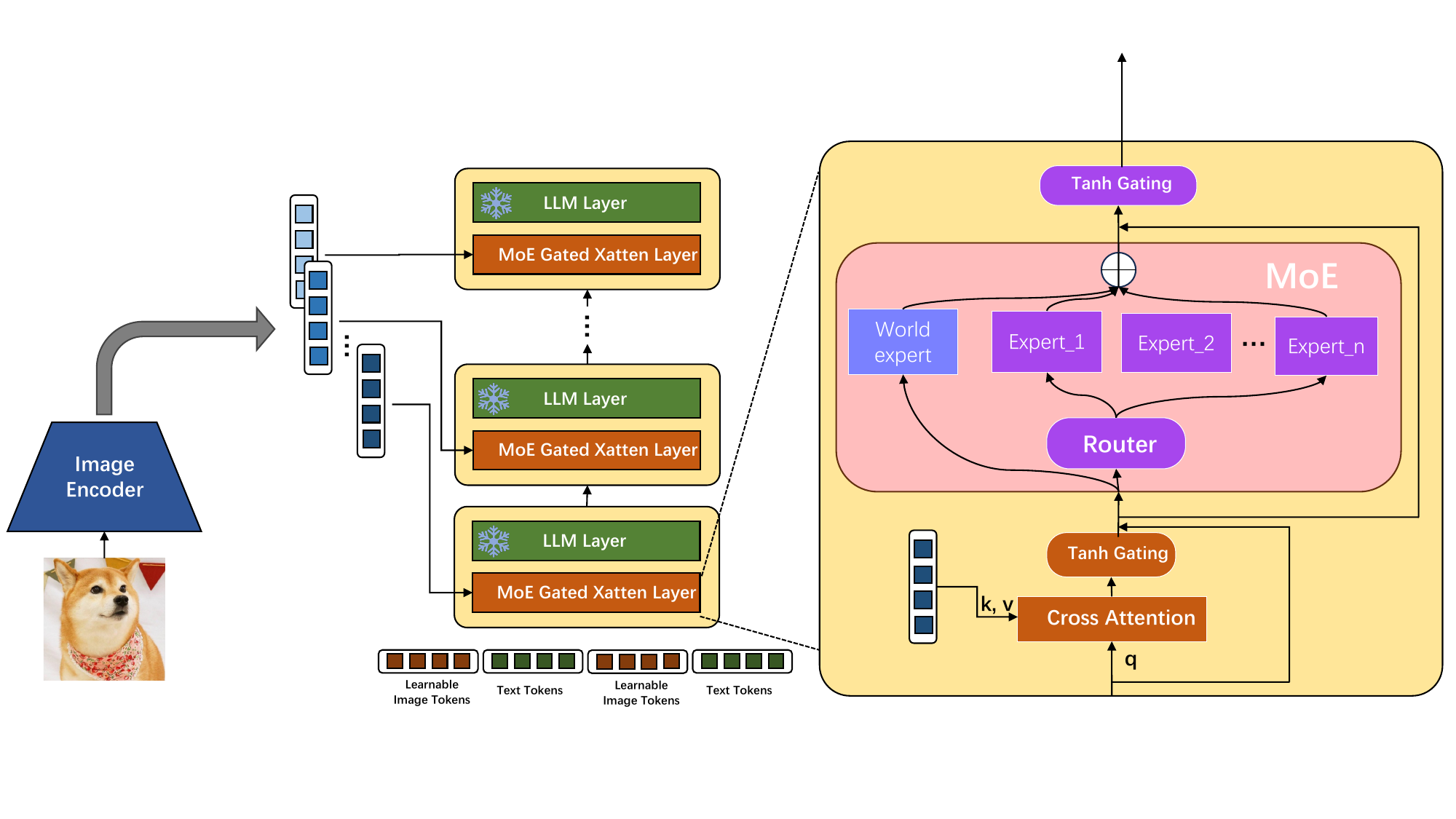}
    \caption{MoE structure.}
\label{moe_model}
\end{figure*}

In order to achieve better performance, we get more training parameters by scaling the Gated Xaaten Layer. As depicted in Fig~\ref{moe_model}, we employ a fine-grained MoE architecture. Initially, we replicate the parameters of the FFN of EVLM-Base $N$ times. Subsequently, each replicated FFN is segmented into $M$ fine-grained experts, resulting in a total of $NM$ fine-grained experts. We choose a routing layer that selects the appropriate set of $k$ fine-grained experts to compute the output for the current token. We have set $n=4$, $m=4$, and $k=4$ in our configuration.

Drawing from established practices\citep{dai2024deepseekmoe}, we introduce the world expert tasked with learning general knowledge. This expert is involved in the processing of every token. The output from the world expert is then combined with the outputs from the fine-grained experts to derive the final result. 

We employ the same training data and configuration of the dense baseline model, and we freeze the LLM and tune only the cross-attention layers and the last quarter ViT layers. The model obtained at this stage is called EVLM-MoE.

\begin{table}[ht]
    \centering
    \caption{Details of supervised fine-tuning data.} 
    \tablestyle{6pt}{1.1}
    \begin{tabular}{ll cc}
         \toprule
         \textbf{Range} & \textbf{Dataset} & \textbf{Type} \\
         \midrule
         \multirow{2}{*}{User Instruct Data} 
         & ShareGPT-4V        & 665K   \\
         & LLaVA-ZH         & 150K  \\
         \midrule
         \multirow{5}{*}{Multimodal Document/Chart Data} & DocVQA  & 10K  \\
         & SynDog-EN & 30K  \\
         & ChartQA    & 18K \\
         & DVQA    & 200K \\
         & AI2D    & 12K \\
         \midrule
         \multirow{3}{*}{Math Problems} & MathInstruct  & 262K  \\
         & MathPlus & 894K  \\
         & geoqa+    & 72K \\
         \midrule
         {Total} &    & 2.3M  \\
         \bottomrule
    \end{tabular}
    \label{tab:pretraining_data}
\end{table}

\section{Evaluation}

In this section, we evaluate various multi-modal tasks to assess our models' visual understanding ability comprehensively.

\subsection{Convergence of Multi-modal Pre-training Stage}

\begin{figure*}[ht]
\centering
\includegraphics[width= 1\textwidth]{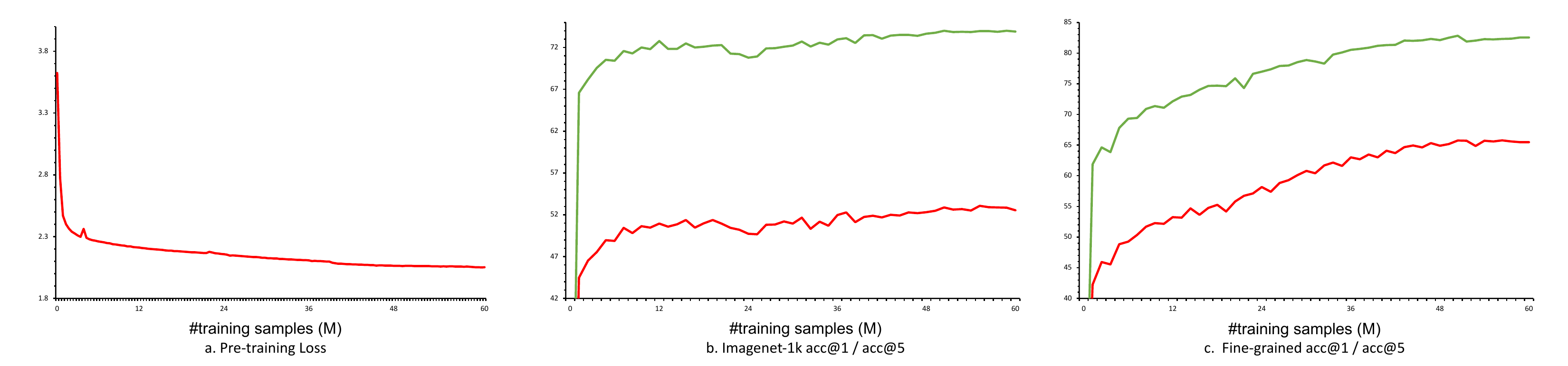}
   \caption{Visualization of the Convergence of the Pre-training Stage}
\label{fig:mllm_stage1}
\end{figure*}

We visualized the convergence of the model during the multi-modal pre-training phase. As shown in Fig.~\ref{fig:mllm_stage1}a, the loss steadily decreases as training progresses. To better monitor the model's alignment between images and text, we randomly sampled 10 examples from each class in the ImageNet-1K~\citep{deng2009imagenet} validation set to assess the model's discriminative ability. During the evaluation, we input a prompt and computed the loss for each of the 1,000 candidate classes, selecting the class with the lowest loss as the model's predicted category to calculate accuracy. From Fig.~\ref{fig:mllm_stage1}b, it can be observed that as training progresses, the recognition accuracy on the ImageNet-1K validation set continues to improve, which serves as an effective monitoring mechanism.

From the evaluation set of ImageNet-1K, we observe rapid convergence of multi-modal large models. This is primarily attributed to the pre-trained parameters of ViT and LLM that we have initialized, enabling effective coarse-grained alignment of multi-modal data with relatively small amounts of image-text pairs. We constructed a finer-grained evaluation set to better monitor the information gain brought by large-scale multi-modal pre-training. This set comprises seven fine-grained categories, including POI, dish, game, and so on. We evaluated these seven categories using methods similar to those used for ImageNet-1K. Fig.~\ref{fig:mllm_stage1}c illustrates the average accuracy across these categories, indicating a steep increase in accuracy for fine-grained recognition as training progresses. This underscores the necessity of large-scale multi-modal pre-training; while coarse-grained alignment is achievable with limited image-text data, a comprehensive understanding of many fine-grained concepts necessitates extensive multi-modal knowledge. The appendix ~\ref{app:fine-grained} presents the variability in accuracy for each of the seven fine-grained categories. Despite extensive pre-training, the accuracy for the "Star" category remains relatively low, suggesting that our current multi-modal pre-training data may not sufficiently cover comprehensive multi-modal knowledge, necessitating further expansion of the dataset scale.

{\color{black}
\subsection{Comparison with State-of-the-Art VLMs}
}

\begin{table*}[t!]
    \scriptsize
    \centering
    \setlength\tabcolsep{0.9pt}
    \renewcommand{\arraystretch}{1.15}
    \caption{\textbf{Comparison with SoTA models on 13 multimodal benchmarks.}General VQA benchmarks include: VQA$^{\text{v2}}$~\cite{vqav2}, GQA~\cite{gqa}, SciQA-Img~\citep{lu2022learn_scienceqa} and VizWiz~\citep{gurari2018vizwiz}.Text-oriented VQA benchmarks include: TextVQA val~\citep{sidorov2020textcaps}, DocVQA \citep{docvqa}, ChartQA~\citep{masry2022chartqa} and AI2D~\citep{kembhavi2016diagram}.General multimodal benchmarks encompass: MME~\cite{fu2023mme}, MMB~\citep{mmbench}, MMB$_{\text{CN}}$~\cite{mmbench} and POPE~\citep{pope}. `*' denotes specialist models obtained from separately fine-tuning on each task.
    }
    
    \begin{tabular}{l|l|l|cccc|cccc|cccc}
        \toprule
            &    &       & \multicolumn{4}{c|}{General VQA} & \multicolumn{4}{c|}{Text-oriented VQA}  & \multicolumn{4}{c}{General Multimodal Benchmarks}\\
    
    \multirow{-2}{*}{Method} & \multirow{-2}{*}{LLM} & \multirow{-2}{*}{Res.} & VQA$^{\text{v2}}$     & GQA       & SciQA-Img       & VizWiz       & TextVQA     & DocVQA           & ChartQA        & AI2D          & MME          & MMB                     & MMB$_{\text{CN}}$           & POPE         	\\ 
    \midrule
    Qwen-VL & Qwen-7B &448$^{\text{2}}$ &79.5 & 59.3 & 67.1 & 35.2 & 63.8 & 65.1 & 65.7 & 62.3 & $-$ & 38.2 & $-$  & $-$ \\
    Qwen-VL-Chat & Qwen-7B & 448$^{\text{2}}$ &78.2 & 57.5 & 68.2 & 38.9 & 61.5 & 62.6 & 66.3 & 57.7 & 1487.58/360.71 & 60.6 & $-$  & $-$ \\
    CogVLM* & Vicuna-7B & 490$^{\text{2}}$ &82.25 & $-$ & 91.0 & $-$ & 70.5 & $-$ & $-$ & $-$ & $-$ & 76.5 & $-$  & 87.88 \\
    LLaVA-1.5 & Vicuna-13B & 336$^{\text{2}}$ &80.0 & 63.3 & 71.6 & 53.6 & 61.3 & $-$ & $-$ & $-$ & 1531/$-$ & 67.7 & 63.6 & 85.9 \\
    InternVL & Vicuna-13B & 336$^{\text{2}}$ &81.2 & 66.6 & $-$ & 58.5 & 61.5 & $-$ & $-$ & $-$ & 1586.4/$-$ & $-$ & $-$ &  87.6 \\
    
    VILA & LLaMA2-13B & 336$^{\text{2}}$ &80.8 & 63.3 & 73.7 & 60.6 & 66.6 & $-$ & $-$ & $-$ & 1570.1 & 70.3 & 64.3  & 84.2 \\
    InfiMM-HD & Vicuna-13B & 448$^{\text{2}}$-1344$^{\text{2}}$ &82.0 & 63.5 & 83.6 & $-$ & 70.7 & 55.1 & $-$ & $-$ & 1472.3/329.4 & 71.6 & $-$  & 87.9 \\
    \rowcolor{gray!15}
    \textbf{EVLM-Base} & Qwen-14B-Chat 1.0 & 448$^{\text{2}}$ & 82.92 & 62.19 & 85.57 & 49.62 & 64.51 & 53.16 & 59.92 & 63.14 & 1579/345 & 78.1 & 71.47  & 94.56 \\
    \rowcolor{gray!15}
    \textbf{EVLM-Chat} & Qwen-14B-Chat 1.0 & 448$^{\text{2}}$ & 81.93 & 64.39 & 86.37 & 47.28 & 67.52 & 53.27 & 63.36 & 76.0 & 1593.56/402.5 & 76.89 & 76.89  & 89.65 \\
    \rowcolor{gray!15}
    \textbf{EVLM-MoE} & Qwen-14B-Chat 1.0 & 448$^{\text{2}}$ & 83.76 & 62.89 & 86.81 & 49.19 & 68.31 & 54.44 & 63.12 & 75.5 & 1607/351 & 78.09 & 76.55  & 93.3 \\
    
    LLava-Next-34B & Yi-34B & 336$^{\text{2}}$*4 &$-$ & $-$ & $-$ & $-$ & $-$ & $-$ & $-$ & 74.9 & 2030.4 & 79.3 & 79  & $-$ \\
    InternVL1.2 & Yi-34B &448$^{\text{2}}$ &$-$ & 64.0 & 83.3 & 60.0 & 72.5 & 57.7 & 68.0 & 79.0 & 1687/489 & 82.2 & 81.2  & 88.0 \\
    CogVLM2 & LLaMA3-Chinese & &$-$ & $-$ & $-$ & $-$ & 85.0 & 88.4 & 74.7 & $-$ & $-$ & 78.9 & $-$  & $-$ \\
    InternVL-1.5 & InternLM2-20B &448$^{\text{2}}$*40 & $-$ & $-$ & $-$ & $-$ & 80.6 & 90.9 & 83.8 & 80.7 & $-$ & 82.2 & 82.0 & $-$ \\
    \bottomrule
    \end{tabular}

    
    \label{tab:sota_results}
    \end{table*}

In this section, we conduct extensive evaluations on a series of benchmarks to assess our model's multimodal understanding and reasoning capabilities. The benchmarks utilized in our study include general VQA, text-oriented VQA, and general Multimodal Benchmarks. As illustrated in Table~\ref{tab:sota_results},  EVLM-Chat and EVLM-MoE demonstrate superior performance compared to its competitors across most of these benchmarks.

\textbf{General VQA Benchmarks.} We utilize four benchmarks: VQA$^{\text{v2}}$, GQA, ScienceQA (Image Set), and VizWiz. For VQA$^{\text{v2}}$, GQA, and VizWiz, we employ a greedy decoding strategy and report the Top-1 accuracy.

Table~\ref{tab:sota_results} presents the overall performance on general VQA tasks. It is crucial to highlight that the evaluations in VQA$^{\text{v2}}$, GQA, and VizWiz are designed to test the models’ visual perception abilities and their capacity to apply prior knowledge effectively. Additionally, ScienceQA, collected from elementary and high school science curricula, contains 21,208 multimodal multiple-choice science questions spanning a wide array of scientific topics, significantly broadening the benchmarking scope.

As shown in Table~\ref{tab:sota_results}, our EVLM-Chat and EVLM-MoE achieve significantly better outcomes than previous generalist models. Specifically, on the ScienceQA task, EVLM-Chat and EVLM-MoE achieved 86.4\% accuracy and 86.8 \% accuracy, respectively. This result even surpasses that of previous generalist models with higher resolution, such as InfiMM-HD, which utilizes a dynamic resolution ranging from 448$^{\text{2}}$ to 1344$^{\text{2}}$. Moreover, our model demonstrates substantial performance improvements in VQA$^{\text{v2}}$, GQA, and VizWiz. These results underscore EVLM's superior capability to integrate multimodal information and utilize extensive prior knowledge for robust reasoning.

\textbf{Text-oriented VQA Benchmarks.} 
In addition to the general VQA evaluation, we further investigate our model's detailed visual perception capabilities by assessing its performance on text-oriented VQA datasets with broad real-world applications. These datasets include TextVQA~\citep{sidorov2020textcaps}, DocVQA~\citep{docvqa}, ChartQA~\citep{masry2022chartqa}, and AI2Diagram~\citep{kembhavi2016diagram}. 

The quantitative results, summarized in Table~\ref{tab:sota_results}, demonstrate that our model outperforms previous general models and recent VLMs on most benchmarks. Notably, on the AI2Diagram dataset, which requires fine-grained visual perception for diagram understanding and associated question answering, EVLM-MoE and EVLM-Chat achieve accuracy of 75.5\% and 76.0\%, respectively. These findings underscore the effectiveness of our proposed deep vision-text fusion in comprehending complex text details within images.
%


\textbf{General Multimodal Benchmarks.} In addition to previous VQA evaluations, we further evaluate our model's visual understanding and reasoning abilities of real-world user behavior on general multimodal benchmarks, including MME, MMB, MMB\(_{\text{CN}}\), and POPE. Compared to traditional VQA datasets, these benchmarks encompass a broader range of evaluation aspects, necessitating more complex reasoning capabilities. 

As summarized in Table~\ref{tab:sota_results}, EVLM-MoE and EVLM-Chat demonstrate commendable overall performance, highlighting its adaptability and capability across various disciplines. Specifically, our model possesses bilingual capabilities due to the large-scale interleaved data of captions, web pages, videos, images, and text. It outperforms previous generalist models on the MMBench and MMBench-Chinese benchmarks. Additionally, our model performs best on the POPE benchmark, showcasing its ability to reduce hallucinations. Our results demonstrate the benefits of vision-language pre-training on downstream tasks. These findings underscore our model's versatility and effectiveness in handling complex visual and textual information.

\subsection{Image Caption}

One of the key capabilities of the Multimodal Large Language Model (MLLM) is DenseCaption of images, which is its most direct application scenario. To enhance MLLM's performance in this area, we integrate high-quality description data into the Supervised Fine-Tuning (SFT) dataset based on existing pre-trained models. This enables the MLLM to generate fluent, detailed, accurate, and illusion-free image descriptions. Given the challenges of annotating DenseCaption, including the high cost and inefficiency of manual rewriting, we have designed a comprehensive process for generating high-quality, detailed image description data.

The process comprises several key steps:

\begin{enumerate}
    \item \textbf{Multiple Descriptions Generation}: Multiple descriptions are generated to ensure comprehensive coverage of all image details.
    \item \textbf{Authenticity Check}: These descriptions are split into short sentences for authenticity verification.
    \item \textbf{Coherent Description Recombination}: The verified sentences are recombined into a coherent description.
    \item \textbf{Stylization Using GPT-4}: Finally, the descriptions are stylized using GPT-4 to ensure they meet specifications and are expressive.
\end{enumerate}

Using the data generated by this process, we effectively guided the MLLM's SFT training, significantly enhancing its image description capability to meet or even exceed human satisfaction. This not only improves MLLM's performance in practical applications but also provides a robust data foundation for future research and development.

\textbf{Auto Caption Pipeline.}
To generate high-quality dense descriptions (DenseCaption), we employ various visual language models (VLMs), including self-developed models, internVL, GPT-4V, and GPT-4o. Initially, these models generate image descriptions which are then split into multiple phrases using the Llama2-70B model.

The split phrases are de-duplicated with the help of Llama2-70B, ensuring each phrase is unique. Subsequently, a powerful multimodal large language model (MLLM), such as GPT-4o, checks the authenticity of these phrases, retaining those that accurately match the image details. Following the authenticity check, GPT-4o integrates these phrases into coherent and fluent image descriptions. This step ensures the resulting descriptions are both accurate and low in illusions.

To further enhance the detail and completeness of the final image descriptions, we integrate multiple MLLM-generated descriptions. This integration improves the quality of the descriptions by making them more detailed and comprehensive. Finally, we use GPT-4 to refine the linguistic expressions, making the generated image descriptions more fluent, elegant, and consistent with human expression preferences.

Through this multi-step, multi-model synergistic approach, we can generate high-quality, detailed, and accurate image descriptions, significantly enhancing the application of multimodal macrolanguage modeling. This method provides robust data support and technical assurance for the development and practical application of multimodal macrolanguage modeling.

\begin{figure*}[ht]
\centering
\includegraphics[width= 1\textwidth]{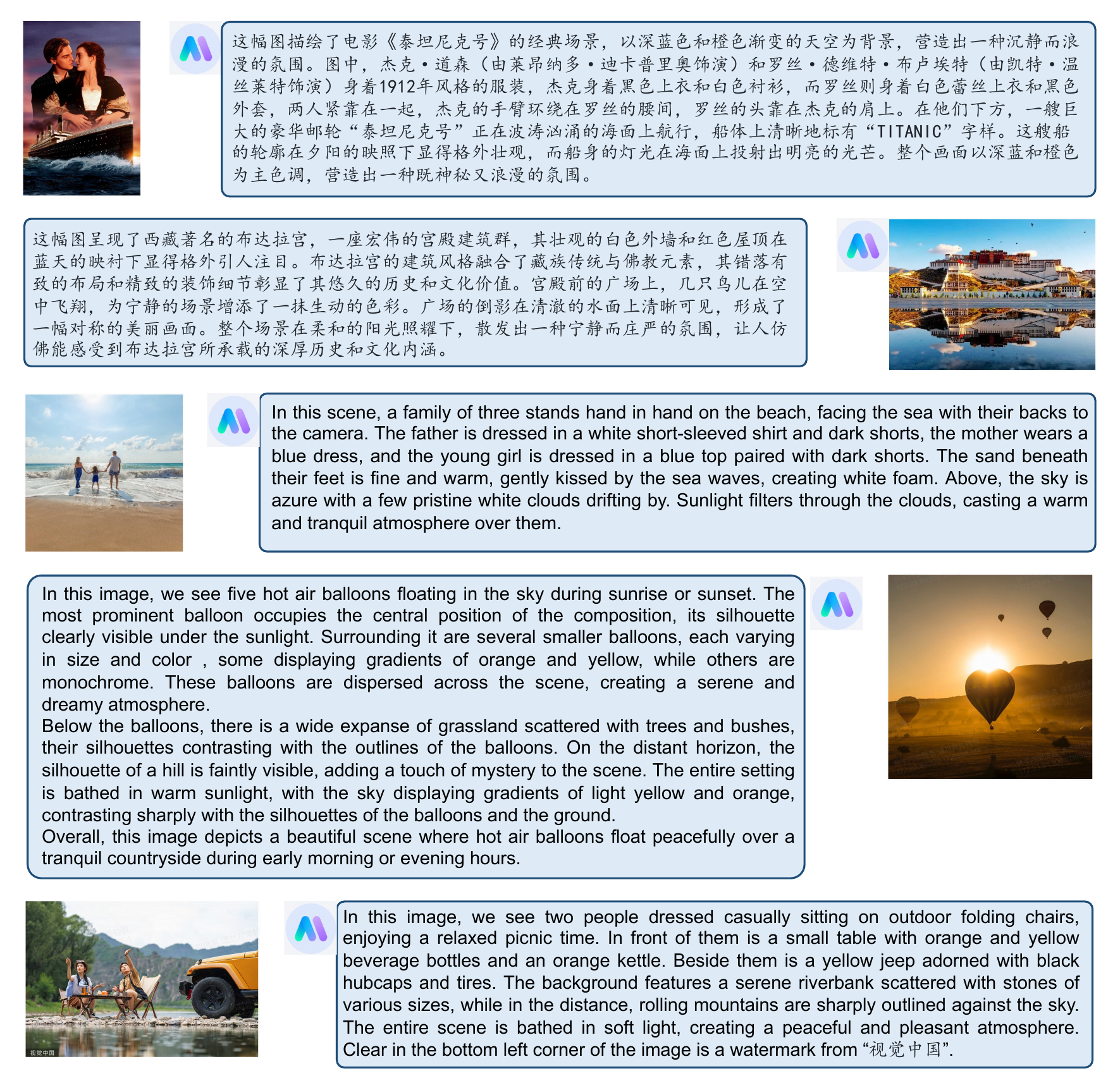}
    \caption{image dense caption.}
\label{mllm_image_caption}
\end{figure*}

Using the image dense descriptions generated by aforementioned process, we fine-tuned our multi-modal large model. Leveraging our model's powerful visual feature perception capabilities, as shown in Fig.~\ref{mllm_image_caption}, we have achieved promising results on the image-dense captioning task. This has notably reduced hallucination phenomena in visual descriptions.

\subsection{Video Caption}
\subsubsection{Attention Mask}
\begin{figure*}[ht]
\centering
\includegraphics[width= 1\textwidth]{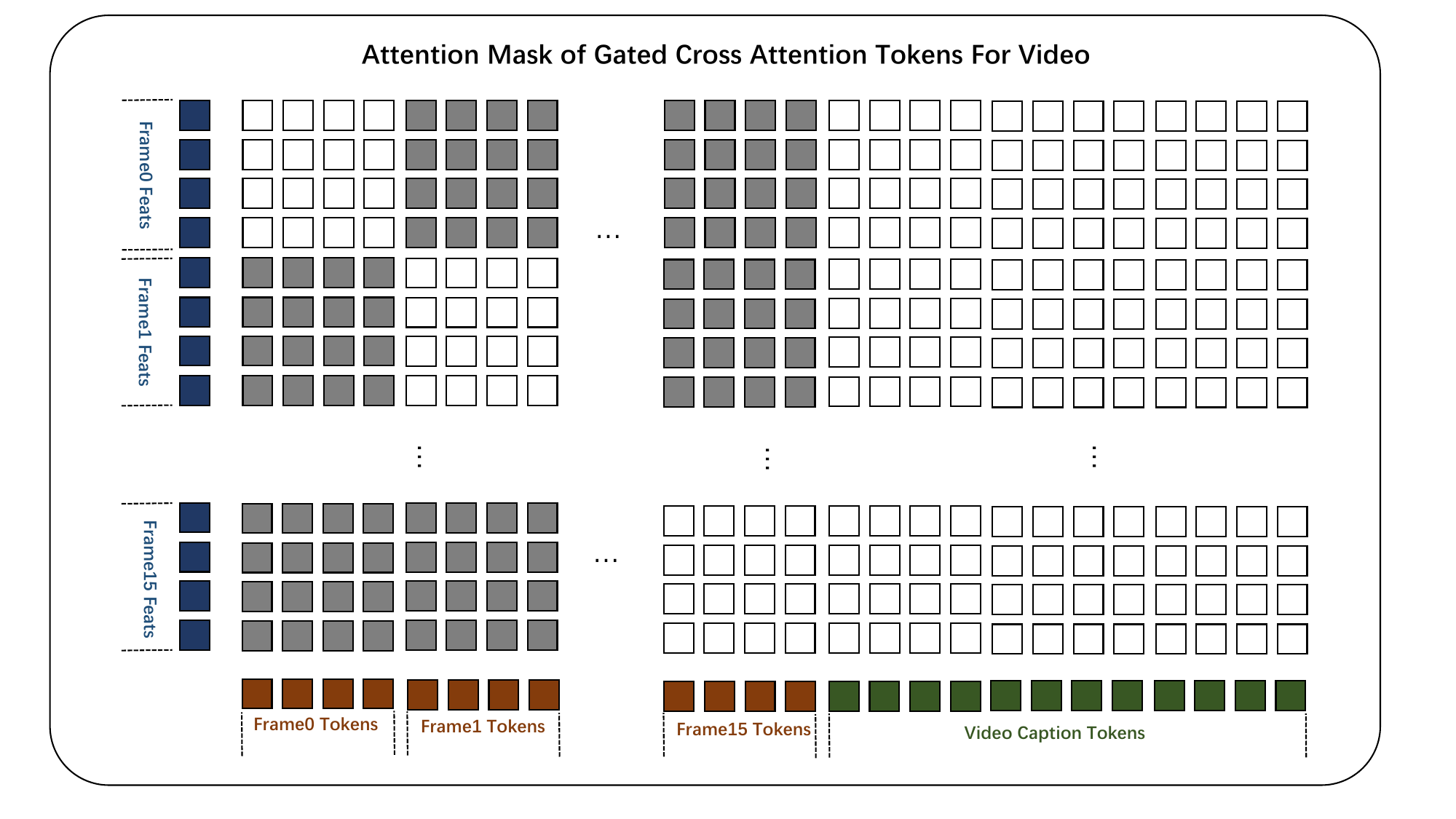}
    \caption{Attention Mask For Video Captioning}
\label{attention_mask_modify}
\end{figure*}

We can also use our EVLM model to understand video. In order to better extract sequence information in the video, such as the action changes of characters in the sequence, the position changes of objects, OCR information in the image, etc., it is necessary to extract information from each image separately when inputting the image sequence into the model to avoid mutual interference between the information of each image. Therefore, it is necessary to design the attention mask in the model during the SFT stage. As shown in Fig.~\ref{attention_mask_modify}, in order to enable the model to acquire all visual information about the video, we ensure that each textual token accesses all visual features related to the video. However, media tokens still only access visual features corresponding to their respective frames.

\subsubsection{Evaluation Benchmarks}
\begin{figure*}[ht]
\centering
\includegraphics[width= 1\textwidth]{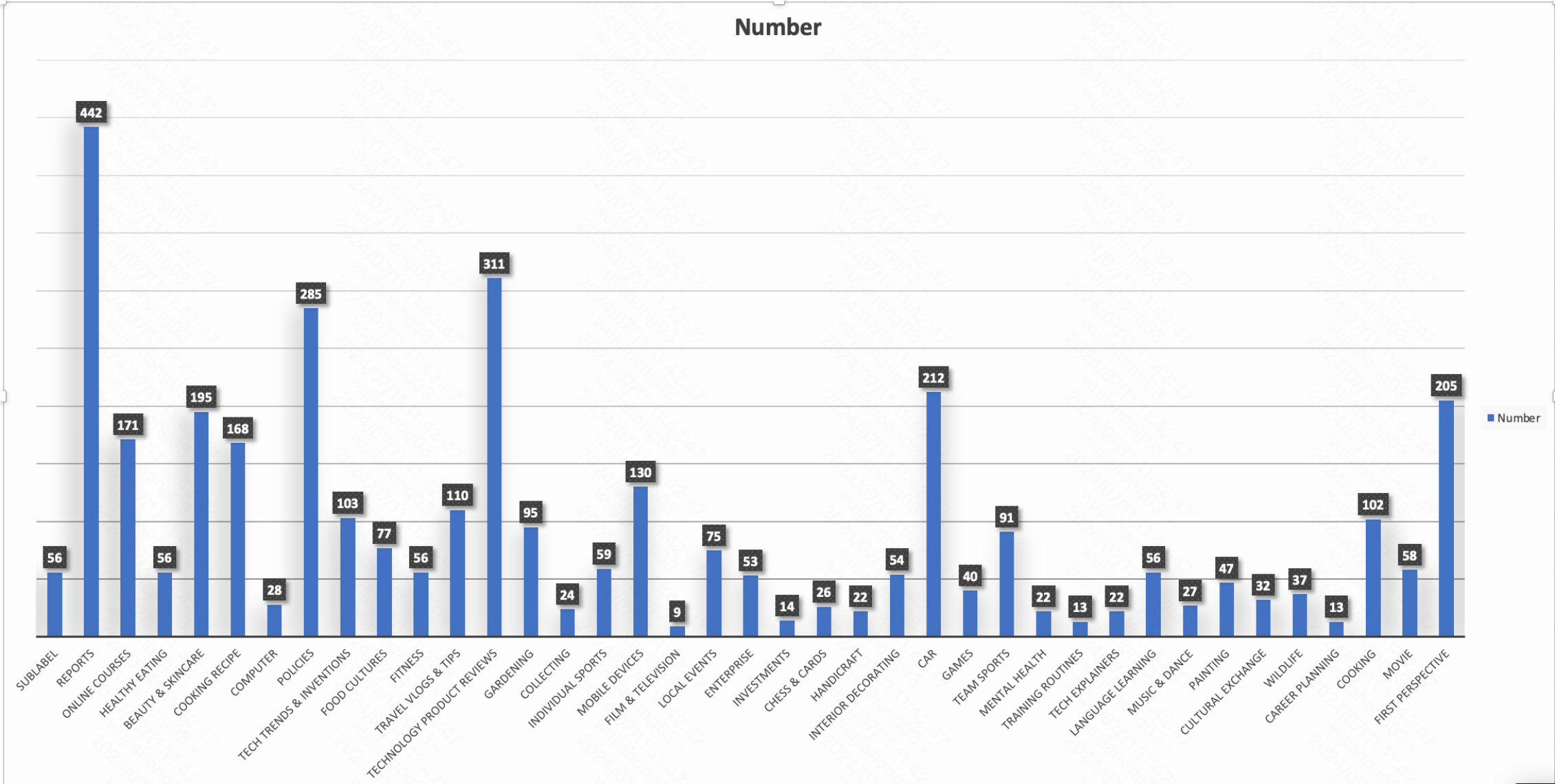}
    \caption{Video caption statistics}
\label{video caption statistics}
\end{figure*}

We use the video-dense caption task to verify the performance of our EVLM model. When constructing the video dense caption, we used five publicly available data sources: YouTube1B ~\citep{zellers2022merlotreserve}, ActivityNet~\citep{Heilbron_2015_CVPR}, Ego4D~\citep{grauman2022ego4dworld3000hours}, with a total of 3596 videos, including 39 categories: Online Courses, Beauty \& Skincare, Computer, etc. As shown in Fig.~\ref{video caption statistics} to meet the diversity of the evaluation benchmark and better test the model's performance. All scores are scored using gpt4o~\citep{achiam2023gpt} as the referee, abandoning the original cider, bleu, and other evaluations, which are more authoritative. The statistical data of each category is shown in the image.
The evaluation results are in the table.
\begin{table}[ht]
    \centering
    \caption{Details of experiments.} 
    \tablestyle{6pt}{1.1}
    \begin{tabular}{ll cc}
         \toprule
         \textbf{Model} & \textbf{Verbosity} & \textbf{Accurate description} \\
         \midrule
         {Video-llava} &  4.06  & 7.0  \\
         {Video-llava2} & 5.69 &7.18\\
         {Video-llama2} & 5.32 &7.1\\
         \boldsymbol{$Ours$} & \boldsymbol{$5.73$} & \boldsymbol{$7.22$} \\
         \bottomrule
    \end{tabular}
    \label{tab:pretraining_data}
\end{table}

\subsubsection{Caption Analysis}
\begin{figure*}[ht]
\centering
\includegraphics[width= 1\textwidth]{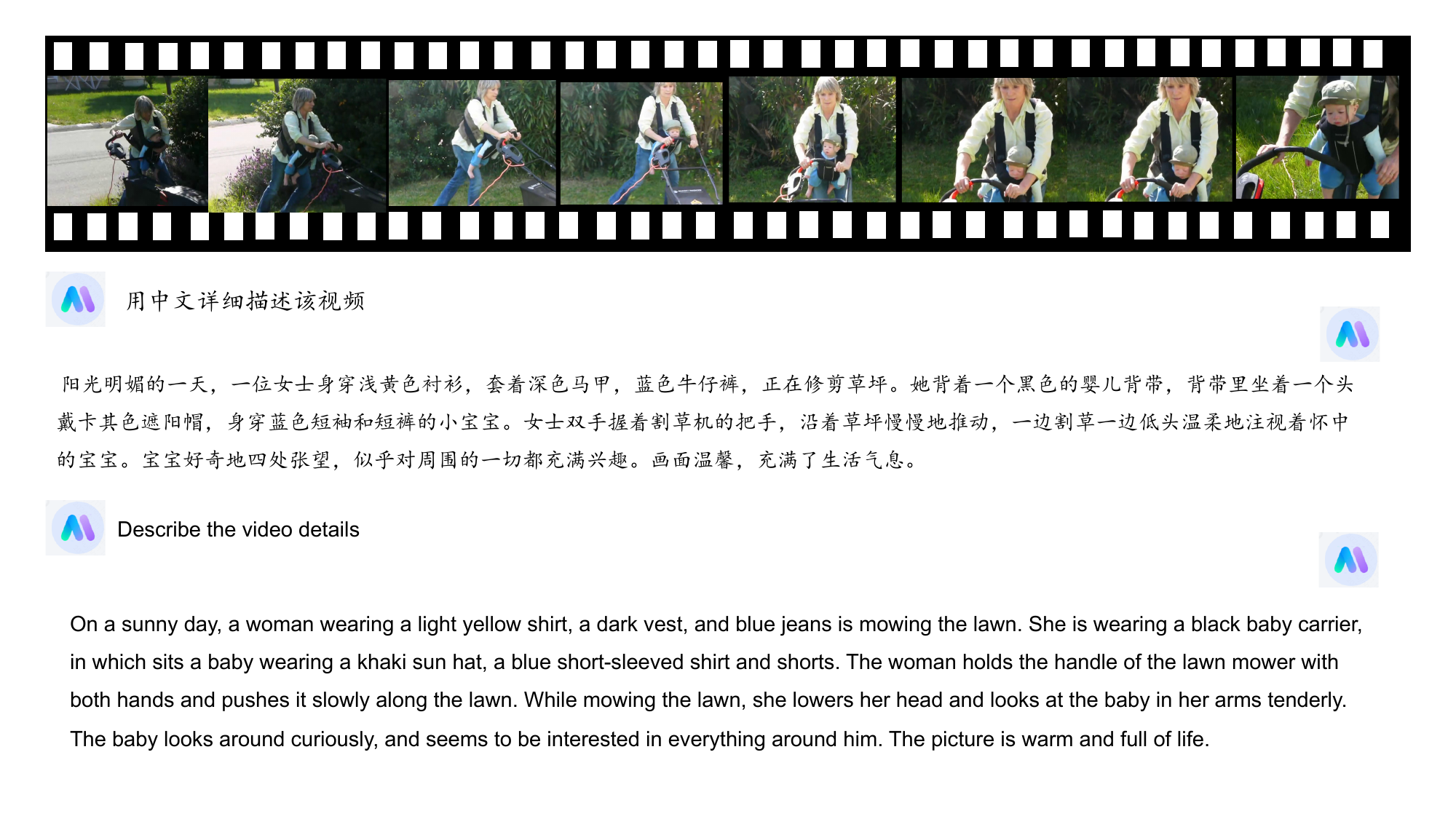}
\label{video caption 2}
\end{figure*}
\begin{figure*}[ht]
\centering
\includegraphics[width= 1\textwidth]{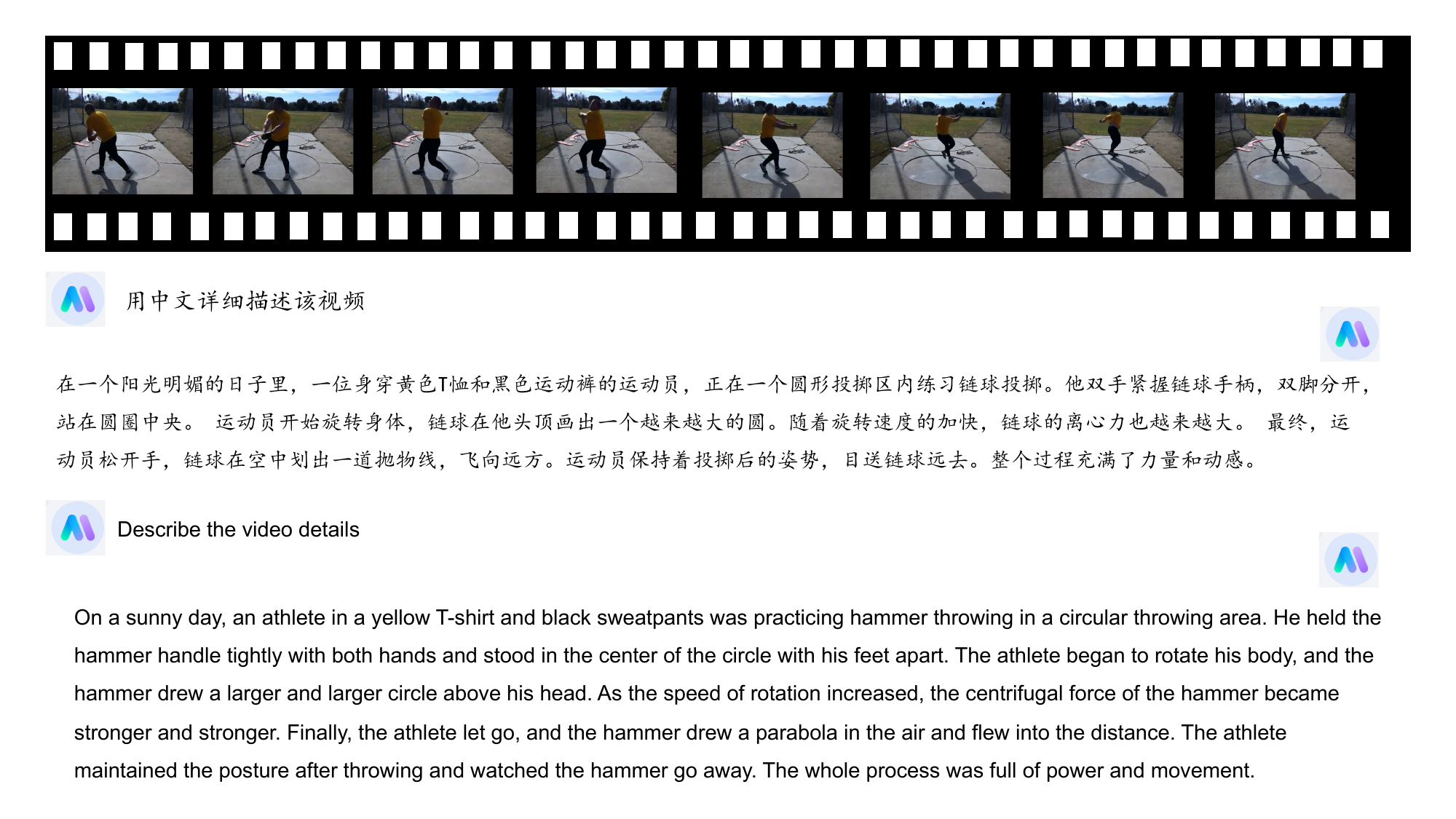}
    \caption{Video caption}
\label{video caption}
\end{figure*}
As shown in figure ~\ref{video caption}, our EVLM large model can generate dense captions in Chinese and English for videos and can well depict the actions and environment of the characters in the video, action categories, and other information.

\section{Related Work}
Recently, multimodal large models have garnered increasing attention, with a multitude of notable works ~\citep{alayrac2022flamingo,blip2,achiam2023gpt,liu2023visual,zhu2023minigpt,dai2023instructblip,bai2023qwen,chen2024far,wang2023cogvlm,kosmos2} emerging in the field. Most of these studies focus on exploring how to more effectively integrate Large Language Models (LLMs) with other modalities to accomplish multimodal tasks. 

\textbf{MLLM Input Project} Most studies employ visual encoders to extract visual features mapped into Large Language Models (LLMs). Some approaches~\cite {liu2023visual,chen2024far} directly feed the output of visual features through a multilayer perceptron (MLP) and concatenate it with the input of the LLM. Another method~\citep {zhu2023minigpt,blip2} adopts a transformer-based structure, commonly referred to as a "q-former," which uses a fixed number of learnable tokens to represent the visual features. Additionally, there are studies~\cite {alayrac2022flamingo,wang2023cogvlm} that integrate visual feature outputs into each layer of the LLM, facilitating a deep fusion of modalities.

\textbf{MLLM Vision Encoder} In the field of multimodal large models, vision models such as CLIP~\citep{clip,openclip}, EvaCLIP~\citep{sun2023eva}, and SigLip~\citep{zhai2023sigmoid} are commonly used as visual encoders. However, to mitigate the potential for information loss inherent in the features extracted by CLIP, some studies~\citep{tong2024eyes} opt to employ an additional vision encoder, such as DINOv2~\citep{oquab2023dinov2}, which is designed to enhance the feature representation. Furthermore, to capture features at varying resolutions and to accommodate the need for efficient computation, some works integrate a lightweight Convolutional Neural Network (CNN)~\citep{he2016deep} model.

\textbf{MoE} The structure of MoE (Mixture of Experts)\cite{jacobs1991adaptive}, characterized by sparse activation, can significantly expand the scale of models or datasets under the same computational resources, thereby enhancing model performance. MoE is widely utilized in LLM and MLLM \cite{fedus2022switch,lin2024moe,dai2024deepseekmoe,jiang2023mixtral}. Upcycling\cite{komatsuzaki2022sparse} proposes to train moe from dense models to reduce training costs. DeepSeek-MoE\cite{dai2024deepseekmoe} enhances the specialization of experts through fine-grained MoE. Additionally, there are studies\cite{mckinzie2024mm1,li2024cumo} applying MoE in MLLMs to further improve model performance.

\vspace{-0.2cm}
\section{Conclusion and Future Work}

We propose an efficient multimodal visual-language model. We can efficiently handle large-scale image-text pre-training by leveraging our refined approach to visual inputs. Our model achieves competitive results on public benchmarks, particularly in image and video-dense captioning. Looking forward, several directions can further enhance model performance:
\begin{itemize}
\item Employing more powerful, larger-scale language models.
\item Exploring the capability of video understanding under extremely long sequences using cross-attention mechanisms.
\end{itemize}

\bibliographystyle{plainnat} 
\bibliography{references}


\clearpage
\newpage

\appendix
\section{Dataset details}
\label{app:dataset}

\subsection{Caption data and Web-type data}
\label{app:data-cleaning-processes}
For the caption dataset, we conducted the following data-cleaning processes:
\begin{enumerate}
\item Removed data containing damaged images and solid color images.
\item Removed data with abnormal aspect ratios.
\item Removed data containing extremely low-resolution images.
\item Removed data with text consisting solely of numbers or symbols.
\item Removed data with text containing long sequences of digits.
\item Removed data where the text contained duplicate words.
\item Removed data containing specific terms such as "HTTP", ".com" and ".png" in the text.
\item Removed data with an excessively short text.
\item Removed data containing date-related text.
\item Converted traditional Chinese characters to simplified Chinese characters.
\item Utilized the CLIP model to calculate image-text relevance and removed data with low relevance scores.
\end{enumerate}

For the web-type dataset, we performed the following straightforward processing steps:
\begin{enumerate}
\item Removed data containing damaged images and solid color images.
\item Removed data with abnormal aspect ratios.
\item Removed data containing extremely low-resolution images.
\item Removed data with an excessive number of images in web data.
\item Removed data where the text length exceeded 2048 characters.
\item Applied relevance filtering similar to MMC4 ~\citep{zhu2024multimodal} to retain highly correlated images.
\end{enumerate}

\subsection{OCR}
\label{app:ocr-data-cleaning-processes}

To enhance the OCR capabilities of our model, we have meticulously curated an OCR dataset sourced from a combination of real-world data and synthetic data. The real-world data we gathered includes content from various sources such as videos uploaded to Kuaishou, the Wukong dataset, Common Crawl 2021, street view data, and a plethora of ebooks presented as images to represent authentic scenarios. To ensure the validity of the image-text pairs at the model's designated resolution, we have implemented the following key steps for processing real-world data:

\begin{enumerate}
    \item Employing expert models to extract texts, coordinate boxes, and recognition confidence prob from the images.
    \item Employing image inpainting techniques to eliminate text with characters that are excessively small or possess low recognition confidence.
    \item Filtering out images with inadequate text content.
    \item Eliminating images that contain redundant text across the entire dataset.
\end{enumerate}

For the synthetic data component, we have harnessed the power of SynthDog to create a diverse OCR dataset, incorporating the use of LaTeX to produce OCR data with dense text. Throughout the data generation process, we begin by selecting text-free images from Kuaishou videos as backgrounds to simulate real-world scenarios. We then explore a wide array of Chinese and English fonts, encompassing both handwritten and standard styles, to generate text in various formats. Furthermore, we introduce uncommon characters, artistic fonts, and diverse data types to enrich the dataset. To bolster the model's ability to recognize dense text, we employ LaTeX to generate PDF data containing a higher volume of characters, subsequently converting them into image-text pairs.

\section{Hyperparameters}
\label{app:hyperparam}
We report the detailed training hyperparameter settings in Table~\ref{tab:hyperparam}.

\begin{table}[htbp]
    \centering
    \tablestyle{7pt}{1.3}
    \caption{Training hyperparameters}
    \begin{tabular}{l ccc}
         \toprule
         Configuration                  & Multi-modal Pre-training & Continual Pre-training & Supervised Fine-tuning \\
         \midrule
         ViT init.                      & EVA2-CLIP-E-PLUS & 1st-stage & 2nd-stage \\
         LLM init.                      & Qwen-14B-Chat 1.0 & Qwen-14B-Chat 1.0 & Qwen-14B-Chat 1.0 \\
         Gated Cross Attention init.    & random & 1st-stage & 2nd-stage \\
         Image resolution         & $224^2$ & $448^2$ & $448^2$ \\
         ViT sequence length      & 257 & 1025 & 1025 \\
         LLM sequence length      & 2048 & 2048 & 2048\\
         Optimizer                & \multicolumn{3}{c}{AdamW} \\
         Optimizer hyperparameter & \multicolumn{3}{c}{$\beta_{1}=0.9, \beta_{2}=0.999, eps=1e^{-8}$} \\
         Peak learning rate       & $6e^{-4}$ & $1e^{-4}$ & $5e^{-5}$ \\
         Minimum learning rate    & $3e^{-5}$ & $5e^{-5}$ & $1e^{-6}$ \\
         ViT Drop path rate       & \multicolumn{3}{c}{0} \\
         Learning rate schedule   & \multicolumn{3}{c}{cosine decay} \\
         Weight decay             & \multicolumn{3}{c}{0.05} \\
         Gradient clip            & \multicolumn{3}{c}{10.0} \\
         Training steps           & 125k & 50k & 12k \\
         Warm-up steps            & 2000 & 2000 & 500 \\
         Global batch size        & 480 & 160 & 16 \\
         Gradient Acc.            & 1 & 1 & 1 \\
         Numerical precision      & \multicolumn{3}{c}{$\mathtt{bfloat16}$} \\
         Data parallel mode       & \multicolumn{3}{c}{FSDP SHARD\_GRAD\_OP} \\
         Activation checkpointing & \multicolumn{3}{c}{\ding{51}} \\
         \bottomrule
    \end{tabular}
    \label{tab:hyperparam}
\end{table}

During the multi-modal pre-training phase, the model was trained using the AdamW optimizer with parameters set as $\beta_{1}=0.9, \beta_{2}=0.999, eps=1e^{-8}$. A cosine learning rate schedule was employed, with a maximum learning rate of $6e^{-4}$ and a minimum of $3e^{-5}$, incorporating a linear warm-up over 2000 steps. We applied a weight decay of $5e^{-2}$ and gradient clipping set to $10.0$. Initially, during the first 25\% of training, only the parameters of the Gated Cross Attention Layer were trained. In the subsequent 75\% phase, the parameters of the latter half of the Visual Encoder were unfrozen for training. The input image size was maintained at $224 \times 224$ pixels throughout this phase. Training encompassed the entire dataset comprising 60 million training samples.

During the continual multi-task training stage, we augmented the input resolution of the visual encoder from $224\times 224$ to $448\times 448$, thereby mitigating information loss associated with image down-sampling. We utilized a cosine learning rate schedule with a maximum learning rate of $1e^{-4}$ and a minimum of $5e^{-5}$, including a linear warm-up over 2000 steps.

\section{Additional experimental details}
\subsection{Convergence of Multi-modal Pre-training Stage}
\label{app:fine-grained}

\begin{figure*}[ht]
\centering
\includegraphics[width= 1\textwidth]{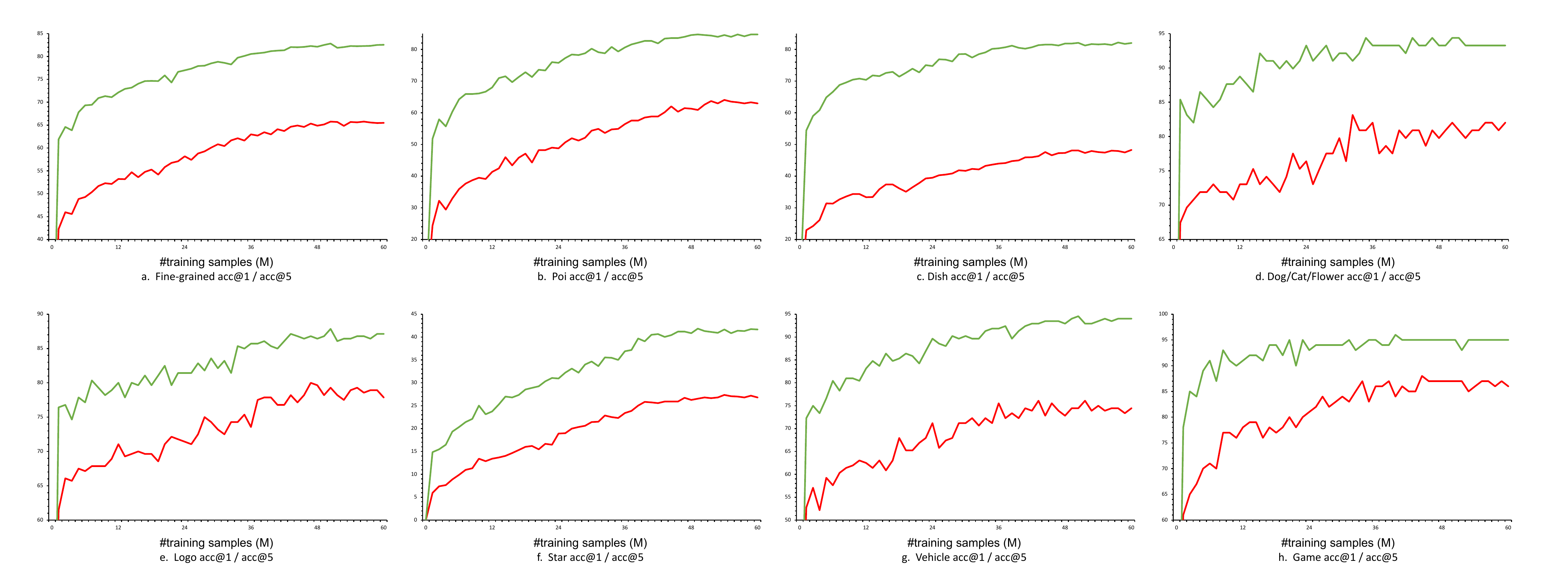}
   \caption{Visualization of the Convergence of the Pre-training Stage}
\label{fig:mllm_stage1_2}
\end{figure*}

Figure \ref{fig:mllm_stage1_2} illustrates the evolution of accuracy across seven fine-grained categories throughout the training process.

\end{document}